\documentclass[10pt,twocolumn,letterpaper]{article}

\usepackage{cvpr}
\usepackage{times}
\usepackage{epsfig}
\usepackage{graphicx}
\usepackage{amsmath}
\usepackage{amssymb}

\usepackage{bbm}
\usepackage{pstricks}
\usepackage{algorithm}
\usepackage[noend]{algpseudocode}
\usepackage{color}
\usepackage{cite}
\usepackage{multirow}
\usepackage{float}

\newtheorem{remark}{Remark}

\usepackage[pagebackref=true,breaklinks=true,letterpaper=true,colorlinks,bookmarks=false]{hyperref}

\cvprfinalcopy 

\def\cvprPaperID{4472} 

\newcommand\blfootnote[1]{%
  \begingroup
  \renewcommand\thefootnote{}\footnote{#1}%
  \addtocounter{footnote}{-1}%
  \endgroup
}

\begin{document}

\title{MTL-NAS: Task-Agnostic Neural Architecture Search towards \\ General-Purpose Multi-Task Learning}

\author{Yuan Gao$^{1*}$, Haoping Bai$^{2*\dagger}$, Zequn Jie$^{1}$, Jiayi Ma$^{3}$, Kui Jia$^{4}$, and Wei Liu$^{1}$ \\
$^1$ Tencent AI Lab ~
$^2$ Carnegie Mellon University  ~ \\
$^3$ Wuhan University ~
$^4$ South China University of Technology \\
{\tt\footnotesize \{ethan.y.gao, bhpfelix, zequn.nus, jyma2010\}@gmail.com, kuijia@scut.edu.cn, wl2223@columbia.edu}
}

\maketitle

\begin{abstract}
   We propose to incorporate neural architecture search (NAS) into general-purpose multi-task learning (GP-MTL). Existing NAS methods typically define different search spaces according to different tasks. In order to adapt to different task combinations (\ie, task sets), we disentangle the GP-MTL networks into single-task backbones (optionally encode the task priors), and a hierarchical and layerwise features sharing/fusing scheme across them. This enables us to design a novel and general \textbf{task-agnostic search space}, which inserts cross-task edges (\ie, feature fusion connections) into fixed single-task network backbones. \\ 
   \indent Moreover, we also propose a novel single-shot gradient-based search algorithm that closes the performance gap between the searched architectures and the final evaluation architecture. This is realized with a minimum entropy regularization on the architecture weights during the search phase, which makes the architecture weights converge to near-discrete values and therefore achieves a single model. As a result, our searched model can be directly used for evaluation without (re-)training from scratch. \\
   \indent We perform extensive experiments using different single-task backbones on various task sets, demonstrating the promising performance obtained by exploiting the hierarchical and layerwise features, as well as the desirable generalizability to different i) task sets and ii) single-task backbones. The code of our paper is available at \url{https://github.com/bhpfelix/MTLNAS}.
\end{abstract}

\section{Introduction}
Recent\blfootnote{* Equal contributions with a random order.}\blfootnote{$\dagger$ Work performed at Tencent AI Lab.} years have witnessed the great success of deep neural networks. It integrates hierarchical feature extraction and optimization in an automatic and end-to-end manner \cite{krizhevsky2012imagenet,simonyan2014very,he2016deep,huang2016densely}. Although deep learning algorithms relieve researchers from feature engineering, they still need carefully designed neural architectures. More recently, Neural Architecture Search (NAS) has received increasing attentions in automating the design of deep neural architectures \cite{Zoph2016NeuralAS}. NAS methods have produced highly competitive architectures on various computer vision tasks, such as image classification \cite{Zoph2016NeuralAS, liu2018darts, Nayman2019XNASNA, Hu2019EfficientFA, pmlr-v97-tan19a, dong2019search, Chen_2019_ICCV, Dong_2019_ICCV}, object detection \cite{Ghiasi_2019_CVPR, chen2019detnas, Xu_2019_ICCV}, and semantic segmentation \cite{Liu_2019_CVPR, Nekrasov_2019_CVPR}.

\begin{figure*}[ht!]
\centering
\vspace{-6mm}
\begin{minipage}{\textwidth}
\centering\includegraphics[width=0.95\textwidth]{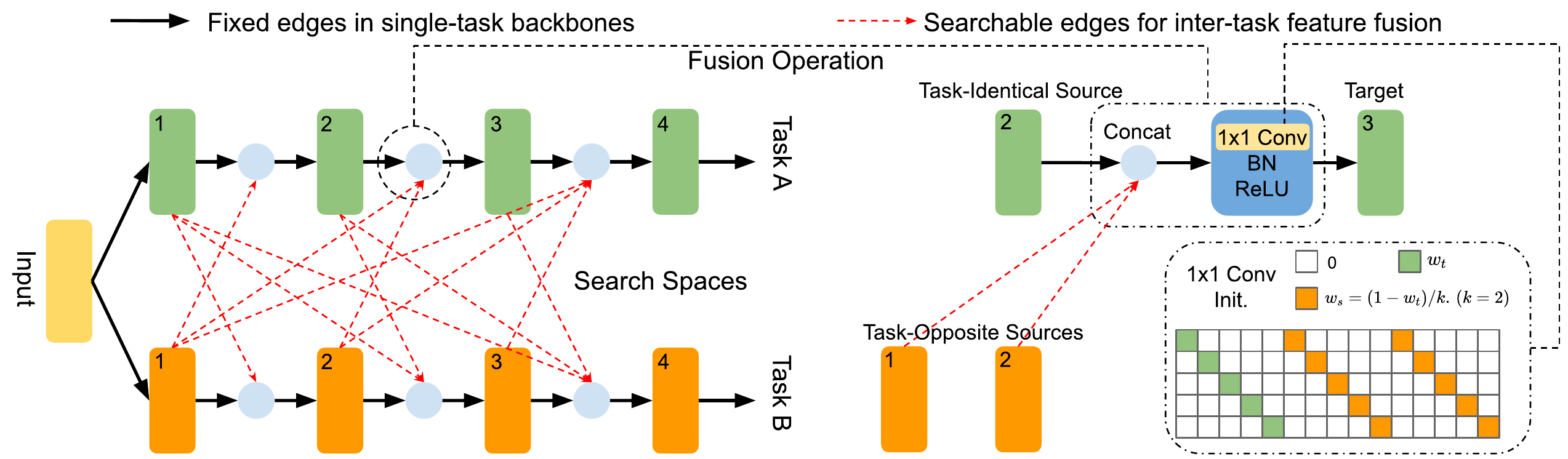}
\end{minipage}
\caption{\label{fig:search_space} The problem formulation of the proposed general-purpose MTL-NAS. We disentangle the GP-MTL networks into fixed task-specific single-task backbones and general feature fusion schemes between them. This allows us to define a general task-agnostic search space compatible with any task combinations, as shown in the leftmost subfigure. The right-top subfigure illustrates the inter-task fusion operation, which is motivated by and extends from the NDDR-CNN \cite{gao2019nddr}. We show the initialization of the fusion operation in the right-bottom subfigure. As we are inserting new edges between the fixed and well-trained single-task network backbones, we wish to make a minimal impact on the original output at each layer at initialization (\ie, initializing with a large $w_t$) (best viewed in color).}
\vspace{-3mm}
\end{figure*}

Another paradigm to boost the performance of the deep neural networks is multi-task learning (MTL) \cite{ruder2017overview,kokkinos2017ubernet}. Multi-task learning has achieved a success in many applications by learning multiple related tasks simultaneously. This success mainly owes to two key factors, \ie, different tasks produce multiple supervision signals that, i) impose additional regularization among different tasks, and ii) produce implicit data augmentation on labels \cite{gao2019nddr,ruder2017overview}. Traditional MTL methods share a single convolutional feature extractor among all tasks, and keep a separate head for each task to produce task-specific predictions. It implicitly assumes that all the learned hierarchical and layerwise features are identical, and are able to perform well on all the tasks. Recent researches show that such an assumption does not always hold \cite{misra2016cross}, \ie, improper feature sharing may introduce negative transfer of some tasks, thus leading to the degradation of the performance \cite{ruder2017overview}.

It is natural to consider incorporating NAS into general-purpose MTL (GP-MTL) to pursue better architectures. To benefit GP-MTL, the NAS algorithm has to adapt to different tasks (or different combinations of tasks, \ie, task sets, in MTL). However, the existing NAS by design prohibits to do so as the search spaces are typically different across different tasks. Fundamentally, those differences in the search spaces reflect different task priors (\eg, see how the prior of semantic segmentation is encoded in Auto-Deeplab \cite{Liu_2019_CVPR}). But when learning on multiple tasks, it is extremely difficult to encode the priors for multiple tasks into the search space, especially when the tasks are loosely related. 

We tackle this difficulty by disentangling the task prior from the feature sharing/fusing scheme in GP-MTL. More specifically, we formulate the GP-MTL paradigm to consist of i) multiple single-task backbone networks (optionally encode the task priors), and ii) the general, hierarchical and layerwise feature sharing/fusing scheme across different backbones (please see the leftmost subfigure of Fig. \ref{fig:search_space}). This formulation enables us to design a general \textbf{task-agnostic search space} for GP-MTL\footnote{Note that only the search space is task-agnostic, the searched architectures are flexible and can be different for different task combinations.}. Specifically, we start with multiple fixed single-task network branches, representing each intermediate layer as a node and the associated feature fusion operations as an edge. The problem thus becomes to seek the optimal edges between the inter-task node pairs, where \emph{the search space is unified for any task set}.

Moreover, we also propose a novel single-shot gradient-based search algorithm that closes the performance gap between search and evaluation. Specifically, it is noticed that the validation performance obtained in the search phase cannot always generalize to evaluation. Fundamentally, it is because that the architecture mixture optimized during the search phase often fails to converge to a discrete architecture, leading to a drop in performance when the final architecture is derived \cite{xie2018snas}. 
We address this issue by reducing the uncertainty of the architecture via entropy minimization during the search phase, resulting in a single model for evaluation directly without the need of (re-)training from scratch, which significantly reduces the performance gap existing in the popular single-shot gradient-based NAS algorithms \cite{liu2018darts,xie2018snas}. 

Note that we focus on GP-MTL in this paper. The proposed method can adapt to different task combinations to generate different inter-task architectures. We fix the single-task backbone branches and search good inter-task edges for hierarchical and layerwise feature fusion/embedding. We also note that the task priors of specific backbone networks (\eg, large convolution kernel in semantic segmentation) can be exploited to further improve the performances on those tasks. But learning the task-specific backbone architecture itself is beyond our scope of GP-MTL. Instead, we design a task-agnostic search space to learn the feature sharing/fusing architecture. We also verify the consistent performance improvement by applying the proposed method to different backbone architectures against the GP-MTL methods \cite{gao2019nddr,misra2016cross}. In summary, our contribution lies in both the search space and the search algorithm:
\begin{itemize}
  \item \textbf{Search Space:} We define a novel task-agnostic search space that enables us to exploit the hierarchical and layerwise feature fusion scheme for GP-MTL, decoupled from the task priors within fixed single-task network backbones. This also makes our method different from the current NAS paradigm, where we are searching/inserting novel inter-task edges into fixed and well-trained network backbones.
  \item \textbf{Search Algorithm:} We propose a novel single-shot gradient-based search algorithm. It alleviates the performance gap between the search phase and the evaluation phase. We achieve this by imposing minimum entropy regularization on the architecture weights. This enables the architecture mixture to converge to a single model/architecture, which will be directly available for evaluation without the need of architecture pruning or re-training from scratch.
\end{itemize}


\section{Related Works}
\noindent \textbf{Neural Architecture Search.} Recently, many neural architecture search (NAS) methods have emerged to jointly learn the weights and the network architectures \cite{Zoph2016NeuralAS, cai2018proxylessnas, liu2018progressive, Chen_2019_CVPR, Nekrasov_2019_CVPR, Li_2019_CVPR, Ghiasi_2019_CVPR, Perez-Rua_2019_CVPR, elsken2018neural,He2020MiLeNAS,mei2020atomnas}. This is realized by various search algorithms including Bayesian optimization \cite{Bergstra:2013:MSM:3042817.3042832}, reinforcement learning \cite{Zoph_2018, Guo_2019_CVPR, Tan_2019_CVPR}, evolutionary algorithms \cite{Real_2019,xie2017genetic}, network transformation \cite{elsken2018simple, Gordon_2018}, and gradient descent \cite{AkimotoICML2019, Wu_2019_CVPR, Zhang_2019_CVPR, Liu_2019_CVPR}. A lot of works employ a single-shot learning strategy (\ie, sharing the model weights and sampling different architectures), which drastically reduces the search time \cite{guo2019single, pham2018efficient, DBLP:conf/nips/SaxenaV16, pmlr-v80-bender18a}. Our work is built on the single-shot gradient-based search methods, where we propose a novel search space and a novel search algorithm tailored to multi-task architecture learning. Our search algorithm is directly related to two popular algorithms in this category, \ie, DARTS \cite{liu2018darts} and SNAS \cite{xie2018snas}. We analyze the objective bias in DARTS and the sampling variance in SNAS, and propose a unified solution, \ie, entropy minimization, to alleviate both the issues. More recently, MTL and multi-modal learning have been exploited using NAS \cite{Perez-Rua_2019_CVPR,liang2018evolutionary}. Our method is different from both of them, \ie, we disentangle the problem into task-specific backbones and general inter-task feature fusion connections, which enables us to design a unified search space to realize \emph{general-purpose} MTL-NAS.

\noindent \textbf{Multi-Task Learning.} Multi-task learning has been integrated with deep neural networks to boost performance by learning multiple tasks simultaneously. A great success has been witnessed in various areas such as detection \cite{girshick2014rich,girshick2015fast,ren2015faster,he2017mask,tdm_arxiv16,primingfeedback_eccv16}, depth prediction and semantic segmentation (also surface normal prediction) \cite{eigen2015predicting,xu2018pad}, human-related tasks \cite{xia2017joint,trottier2017multi,han2017heterogeneous,ranjan2017hyperface,yim2015rotating}, \emph{etc}. Our work leverages NAS to move towards GP-MTL, which (or whose search space) is compatible with any task combinations \cite{long2015learning,kokkinos2017ubernet,lu2016fully,yang2016deep,kendall2017multi,misra2016cross,ruder2019latent,liu2019end,maninis2019attentive}. Our method (or more specifically, our search space design) is mainly inspired by recent researches including cross-stitch networks \cite{misra2016cross} and NDDR-CNN \cite{gao2019nddr}, which enables to treat GP-MTL as an inter-task feature fusion scheme decoupled from the task-specific backbone networks. Our method extends \cite{misra2016cross,gao2019nddr} by allowing fusion of an arbitrary number of source features. Moreover, rather than heuristically inserting the fusion operation into selected layers in \cite{misra2016cross,gao2019nddr}, our method can automatically learn better fusion positions.


\section{Problem Formulation} \label{sec:search_space}

We introduce our problem formulation in this section. Firstly, we show how to dissect the GP-MTL problem by disentangling the task-specific backbone networks and general inter-task feature fusion in Sect. \ref{sec:disentangle}. In Sect. \ref{sec:search_space}, we formally present our task-agnostic search space, which is unified for different task combinations. Finally, we detail our choice of feature fusion operation in Sect. \ref{sec:feature_fuse_op}.


\subsection{Task-Specific Backbone Networks and General Inter-Task Feature Fusion \label{sec:disentangle}}
Arguably, the main difficulty to integrate NAS into GP-MTL is the difference in the designed search spaces for different tasks. This is because the search space by definition should reflect the inductive bias of the associated task. The situation is even more rigorous for GP-MTL as we have exponentially more task combinations (\ie, task sets).

Our strategy is to disentangle multi-task architecture into a (general) shared structure involving feature fusion connections and the task-specific sub-networks, and optimize the shared structure with NAS. Recent researches in GP-MTL \cite{misra2016cross,gao2019nddr} inspire us to formulate the single-task backbones as the task-specific parts, while focusing on designing the feature fusion scheme with a general \emph{task-agnostic} search space which is independent of different task combinations. We illustrate the fixed single-task backbone networks and the learnable inter-task fusion edges in the leftmost subfigure of Fig. \ref{fig:search_space}. 

\begin{figure*}[t]
\vspace*{-10pt}
\centering
\normalsize
\setcounter{equation}{2}
\begin{equation}
\label{eq:fusion_op}
O_j = \texttt{ReLU} \Big(\texttt{BN} \big( [w_{\text{TI}}\mathbb{I}, \, \overbrace{w_{\text{TO}}\mathbb{I},  ..., w_{\text{TO}}\mathbb{I}}^\text{$j$ elements}] [F_{j}^{\text{TI}}, \, z_{0j}\texttt{R}(F_0^{\text{TO}}), ..., z_{jj}\texttt{R}(F_j^{\text{TO}})]^\top \big) \Big) = \texttt{ReLU} \Big(\texttt{BN} \big( [w_{\text{TI}} F_{j}^t + w_{\text{TO}} \sum_{i=1}^{j} z_{jj}\texttt{R}(F_j^{\text{TO}})] \big) \Big)
\end{equation}
\hrulefill
\vspace*{-7pt}
\setcounter{equation}{0}
\end{figure*}

\subsection{Search Space \label{sec:search_space}}
We formally describe the task-agnostic search space for GP-MTL based on the above discussion. We consider the same GP-MTL scenario as described by in \cite{misra2016cross}, where two tasks A and B share the same input. Our goal is to construct a multi-task network by learning good inter-task feature fusion edges on two well-trained single-task networks.

We aim to search a direct-acyclic graph (DAG) by adding directed edges at every intermediate layer of each fixed single-task network (please see the leftmost subfigure of Fig. \ref{fig:search_space}). Every directed edge (\ie, computations) points from the \emph{source} features to the \emph{target} features. We have two types of source features in the GP-MTL framework. Considering learning on Task A, we denote the source features from the same task A as the \emph{task-identical source} features. The source features from the other task(s) are \emph{task-opposite source} features, which provide complementary features from the opposite task. We fix the \emph{task-identical source} edge and search over the complete set of possible \emph{task-opposite source} edges. 

Formally, we aim to calculate the optimal fused feature at the $j^{th}$ \emph{target} layer $O_j$, exploiting the \emph{task-identical source} features $F_j^{\text{TI}}$ and the completed set of the candidate \emph{task-opposite source} features $\mathcal{S}^{\text{TO}}$. The construction of $\mathcal{S}^{\text{TO}}$ determines the extent of our search space. In order to avoid creating cycles in the resulting multi-task architecture, we limit the indices of the candidate \emph{task-opposite source} features to be not larger than $j$. We denote such a limited candidate \emph{task-opposite source} feature set as $\mathcal{S}^{\text{TO}}_j = [F_0^{\text{TO}}, \cdots, F_j^{\text{TO}}]$. Therefore, our search space associated with the \emph{task-identical source} feature $F_j^{\text{TI}}$ can be characterized by a tuple $(\mathcal{S}^{\text{TO}}_j, \mathcal{C})$, where $\mathcal{C}$ is the fusion operations on $F_j^{\text{TI}}$ and $\mathcal{S}^{\text{TO}}_j$. Finally, the optimal fused feature $O_j$ is:
\begin{equation}
\label{eq:final_search_space}
\scalebox{0.95}{$
O_j = \mathcal{C}(F_j^{\text{TI}}, \mathcal{S}^{\text{TO}}_j) = G\Big(H\big( [F_j^{\text{TI}}, z_{0j}\texttt{R}(F_0^{\text{TO}}), ..., z_{jj}\texttt{R}(F_j^{\text{TO}})] \big)\Big), 
$}
\end{equation}
where $\mathcal{C} = \{G, H\}$ with $G$ being the nonlinear activation and $H$ being the feature transformations. \texttt{R} is the spatial resizing operation (\eg, bilinear interpolation) to enable concatenation. Each $z_{i j}$ is a binary indicator denoting if there is an edge from the $i$-th task-opposite source node to the $j$-th target node, which will be optimized by the NAS algorithm.

For two tasks with $n$ layers for each, this general search space can produce $2^{n(n+1)}$ candidate fusion architectures, including the state-of-the-art NDDR-CNN \cite{gao2019nddr} and cross-stitch networks \cite{misra2016cross} as the special cases.





\subsection{Inter-Task Feature Fusion Operation \label{sec:feature_fuse_op}} 
We follow NDDR-CNN \cite{gao2019nddr} to design our feature transformation $H = \{\texttt{1x1 Conv}(\cdot) \}$ and our nonlinear activation $G = \{\texttt{ReLU} \big(\texttt{BN} (\cdot) \big) \}$. Note that our fusion operation in Eq. \eqref{eq:final_search_space} generalizes that of NDDR-CNN as we can take an \emph{arbitrary} number of input features, which enables the convergence to a \emph{heterogeneous and asymmetric} architecture\footnote{Note that it is easy to incorporate more candidate operations into $H$ and $G$. We have tested including summation into $H$ but witnessed a negligible improvement. Therefore, we fix the design of $H$ and $G$ for simplicity.}. Our fusion operation in Eq. \eqref{eq:final_search_space} becomes the following Eq. \eqref{eq:actual_fusion}, which is also shown in the right-top subfigure of Fig. \ref{fig:search_space}:
\begin{equation}
\scalebox{0.95}{$
O_j = \texttt{ReLU} \Big(\texttt{BN} \big (\texttt{1x1Conv} [F_j^{\text{TI}}, z_{0j}\texttt{R}(F_0^{\text{TO}}), ..., z_{jj}\texttt{R}(F_j^{\text{TO}})] \big) \Big). $}
\label{eq:actual_fusion}
\end{equation}

We also note that the initialization of $H$ (\ie, the $1 \times 1$ convolution) is important, as we are inserting novel edges into fixed and well-trained single-task backbones. Therefore, we should avoid severe changes of the original single-task output at each layer. Formally, we show the initialization for the operation in Eq. \eqref{eq:fusion_op}.

In Eq. \eqref{eq:fusion_op}, $\mathbb{I}$ is an identity matrix which enables us to focus on only initializing the block-diagonal elements in the $1 \times 1$ convolution, $w_\texttt{TI}$ and $w_\texttt{TO}$ are the initialized weights for the \emph{task-identical source} and the \emph{task-opposite source} features, respectively. We empirically set $w_\texttt{TI} + j w_\texttt{TO} = 1$ and initialize a large $w_\texttt{TI}$ similar to \cite{gao2019nddr,misra2016cross}. 
The initialization is illustrated in the right-bottom subfigure of Fig. \ref{fig:search_space}.

\section{Search Algorithm}
In this section, we present our single-shot gradient-based search algorithm, which optimizes the model weights and the architecture weights over the meta-network (\ie, the network includes all legal connections defined by the search space) by gradient descent. Our method is able to alleviate the performance gap between the searched and the evaluation architectures, where the performance gap was caused by the inconsistency between the searched mixture architectures and the derived single evaluation architectures in previous single-shot gradient-based search algorithms.

Fundamentally, the undesirable inconsistency in the searched and evaluation architectures is introduced by the \textbf{continuous relaxation} and the \textbf{discretization} procedures of the single-shot gradient-based search algorithms. In order to better understand this, we first discuss the \emph{continuous relaxation} and the \emph{discretization} in Sect. \ref{sec:relaxation}. Based on that, in Sect. \ref{sec:variance}, we analyze the objective bias caused by the inconsistency between the deterministic continuous relaxation and deterministic discretization. We note that stochastic SNAS \cite{xie2018snas} was proposed to solve the objective bias, but it may introduce a large sampling variance, leading to an unstable evaluation performance. Finally, we propose the minimum entropy regularization to alleviate both the issues and present our optimization procedure in Sect. \ref{sec:optimization}.

\subsection{Continuous Relaxation and Discretization} \label{sec:relaxation}
Typical single-shot gradient-based methods usually contain two stages, \ie, i) continuous relaxation and ii) discretization. The continuous relaxation enables the gradients to be calculated and back-propagated to search the architecture, as the original objective of the NAS methods is discrete and non-differentiable. As the search phase often converges to a mixture model (with many architecture weights between 0 and 1), we thus need to derive a single child network for evaluation by discretization.

We denote the connectivity of the network $\boldsymbol{Z}$ as a set of random variables $Z_{ij}$, where $Z_{ij}$ is sampled from some discrete distribution parameterized by the \emph{architecture weights} $\alpha_{ij}$, \ie, $\boldsymbol{Z} = \{Z_{ij} \ \sim \texttt{DisDist}(\alpha_{ij}) \ | \ \forall (i, j) \ \text{in the search space}\}$, where $i, j$ refer to the source node (\ie, the input position) and the target node (\ie, the output position to perform the associated operations), respectively. Here, the discrete sampling distribution can be \texttt{Categorical} or \texttt{Bernoulli}, depending on whether there are multiple or only one candidate operations to search. We use a Bernoulli distribution to present our problem as there is only the NDDR feature fusion operation in our search space (see Sect. \ref{sec:feature_fuse_op} and Eq. \eqref{eq:actual_fusion}), but note that the proposed search algorithm is general and can also be used with multiple candidate operations.

We denote the multivariate sampling distribution for all the fusion connections as $p_{\boldsymbol{\alpha}} (\cdot)$, where $\boldsymbol{\alpha} = \{\alpha_{ij} \ | \ \forall (i, j) \ \text{in the search space}\}$. The architecture search objective is \cite{pmlr-v80-pham18a,xie2018snas}:
\begin{equation}
\setcounter{equation}{3}
\label{eq:objective}
\underset{\boldsymbol{\alpha}}{\text{min}} \ \mathbb{E}_{\boldsymbol{Z} \sim p_{\boldsymbol{\alpha}}(\boldsymbol{Z})} [\mathcal{L}_{\theta}(\boldsymbol{Z})],
\end{equation}
where $\theta$ is a set of CNN weights and $\mathcal{L}_{\theta}(\boldsymbol{Z})$ is the loss function of the sampled (discrete) architecture $\boldsymbol{Z}$.

In order to optimize $\boldsymbol{\alpha}$ with gradient based methods, one solution is to relax the discrete sampling procedure $\boldsymbol{Z} \sim p_{\boldsymbol{\alpha}}(\boldsymbol{Z})$ to be continuous:

\noindent \textbf{Deterministic Relaxation.} The deterministic approaches such as DARTS directly maintain and optimize a mixture of architectures. For Bernoulli random variable $Z_{ij} \sim \texttt{Ber}(\alpha_{ij})$, it directly uses the distribution mean instead of the discrete samples, \ie, $Z_{ij} = \alpha_{ij}$. Therefore, the relaxed objective of Eq. \eqref{eq:objective} and fusion operation of Eq. \eqref{eq:final_search_space} become:
\begin{align}
& \underset{{\color{red} \boldsymbol{\alpha}}}{\text{min}} \ \mathcal{L}_{\theta}({\color{red} \boldsymbol{\alpha}}) \label{eq:darts_obj}, \\
& O_j = G\Big(H\big( [F_j^{\text{TI}}, {\color{red} \alpha_{0j}} \texttt{R}(F_0^{\text{TO}}), ..., {\color{red} \alpha_{jj}} \texttt{R}(F_j^{\text{TO}})] \big)\Big) \label{eq:darts}.
\end{align}


\noindent \textbf{Stochastic Relaxation.} SNAS \cite{xie2018snas} uses the reparameterization trick with the \texttt{Concrete} distribution \cite{Maddison2016TheCD} to sample architectures during search while allowing gradient to back-propagate through the sampling procedure. The reparameterized multivariate \texttt{Bernoulli} is: 
\begin{equation}
\label{eq:concrete}
\boldsymbol{X}  = \frac{1}{1 + \exp(-(\log(\boldsymbol{\alpha}) + \boldsymbol{L})/\tau)} \sim q_{\boldsymbol{\alpha}}(\boldsymbol{X}),
\end{equation}
where $\boldsymbol{X}$ is the continuous relaxation of $\boldsymbol{Z}$, \ie, each entry in $\boldsymbol{X}$ takes a continuous value in $[0, 1]$, and $\boldsymbol{L} = (L_1, L_2, \cdots, L_n)$ with $L_i \sim \texttt{Logistic}(0, 1)$. As the temperature $\tau$ approaches 0, each element in $\boldsymbol{X}$ smoothly approaches a discrete \texttt{Binomial} random variable.

As a result, the relaxed objective of Eq. \eqref{eq:objective} and the fusion operation of Eq. \eqref{eq:final_search_space} become:
\begin{align}
& \underset{{\color{red} \boldsymbol{\alpha}}}{\text{min}} \ \mathbb{E}_{{\color{red} \boldsymbol{X} \sim q_{\boldsymbol{\alpha}}(\boldsymbol{X})}} [\mathcal{L}_{\theta}({\color{red} \boldsymbol{X}})], \label{eq:snas_obj} \\
& O_j = G\Big(H\big( [F_j^{\text{TI}}, {\color{red} x_{0j}} \texttt{R}(F_0^{\text{TO}}), ..., {\color{red} x_{jj}} \texttt{R}(F_j^{\text{TO}})] \big)\Big), \label{eq:reparameterized_computation}
\end{align}
where $x_{ij}$ is the sampled connectivity from source node $i$ to target node $j$, \ie, $\boldsymbol{X} = \{x_{ij} \ | \ \forall (i, j) \ \text{in the search space} \}$.




The search optimizations on both the relaxed objectives \emph{converge to a mixture model with $\alpha_{ij} \in [0, 1]$}, which require discretizations to derive a single child model for evaluation:


\noindent \textbf{Deterministic Discretization.} This is the discretization method used in DARTS, which simply keeps the connection with the highest architecture weight. For our binary case:
\begin{equation}
\texttt{Ind}(\alpha_{ij}) = 
\left\{
\begin{array}{l}
1, \quad \text{if $\alpha_{ij} > 0.5$}, \\
0, \quad \text{otherwise,}
\end{array}
\right.
\end{equation}
where $\texttt{Ind}(\cdot)$ is an indicator function. 

Therefore, the final discretizated child network is:
\begin{equation}
\scalebox{0.95}{$
O_j = G\Big(H\big( [F_j^{\text{TI}}, {\color{red} \texttt{Ind}(\alpha_{0j})} \texttt{R}(F_0^{\text{TO}}), ..., {\color{red} \texttt{Ind}(\alpha_{jj})} \texttt{R}(F_j^{\text{TO}})] \big)\Big).
$} 
\label{eq:darts_redis}
\end{equation}

\noindent \textbf{Stochastic Discretization.} The discretization of SNAS has already taken place during the search optimization Eq. \eqref{eq:snas_obj}. After the search converges, SNAS samples a child architecture of each $x_{ij}$ according to Eq. \eqref{eq:concrete} with the converged $\boldsymbol{\alpha}$ and $\tau=0$, resulting in the same form as Eq. \eqref{eq:reparameterized_computation}.

\subsection{Objective Bias and Sampling Variance} \label{sec:variance}




\noindent \textbf{Objective Bias of the Deterministic Method.} The inconsistency between Eqs. \eqref{eq:darts} and \eqref{eq:darts_redis} introduces an objective bias between the relaxed parent and the discretized child:

\begin{equation}
   \Big\lvert \mathcal{L}_{\theta}(\boldsymbol{\alpha}) - \mathcal{L}_{\theta}(\texttt{Ind}(\boldsymbol{\alpha})) \Big\rvert \geq 0, \label{eq:obj_diff}
\end{equation}
where $\mathcal{L}_{\theta}(\boldsymbol{\alpha})$ is the objective for the search optimization, and $\mathcal{L}_{\theta}(\texttt{Ind}(\boldsymbol{\alpha}))$ is the \emph{true} objective we aim to minimize with the actual \emph{evaluation architecture} $\texttt{Ind}(\boldsymbol{\alpha})$. 

\begin{remark}
Due to the complex and architecture-dependent nature of $\mathcal{L}_{\theta}$, it is difficult to deduce all cases where $\mathcal{L}_{\theta}(\boldsymbol{\alpha}) = \mathcal{L}_{\theta}(\mathtt{Ind}(\boldsymbol{\alpha}))$. Instead, with a well-defined $\mathcal{L}_{\theta}$, \ie, $x = y$ $\implies \mathcal{L}_{\theta}(x) = \mathcal{L}_{\theta}(y)$ and the local Lipschitz continuity near $\mathtt{Ind}(\boldsymbol{\alpha})$, we can simply force $\boldsymbol{\alpha}$ close to $0$ or $1$, therefore achieving $\boldsymbol{\alpha} = \mathtt{Ind}(\boldsymbol{\alpha})$ and ultimately $\mathcal{L}_{\theta}(\boldsymbol{\alpha}) = \mathcal{L}_{\theta}(\mathtt{Ind}(\boldsymbol{\alpha}))$. We achieve this by applying minimum entropy regularization on $\boldsymbol{\alpha}$.
\end{remark}

\noindent \textbf{Sampling Variance of the Stochastic Method.} There is no objective bias in the stochastic method when $\tau = 0$, as the training and evaluation objectives align to Eq. \eqref{eq:reparameterized_computation}\footnote{Note that the bias exists during the search phase when $\tau$ is not 0 \cite{Maddison2016TheCD}.}. However, there does exist a (probably large) variance when sampling the child variance after convergence.

As shown in the \texttt{Concrete} distribution \cite{Maddison2016TheCD}, when the temperature $\tau$ is annealed to 0, $x_{ij}$ is sampled to 1 with probability $\alpha_{ij}$, \ie, $\underset{\tau \rightarrow 0}{\text{lim}} P(x_{ij} = 1) = \alpha_{ij}$. Note that $\alpha_{ij}$ is not well regularized during the search optimization. When $\alpha_{ij}$ converges to 0.5, the sampling of $x_{ij}$ randomly picks 0 or 1, which leads to an unstable evaluation. Our empirical results in Fig. \ref{fig:histo_alpha} show that this can happen in practice.
\begin{figure}[ht!]
\vspace{-3.5mm}
\centering
\includegraphics[width=\linewidth]{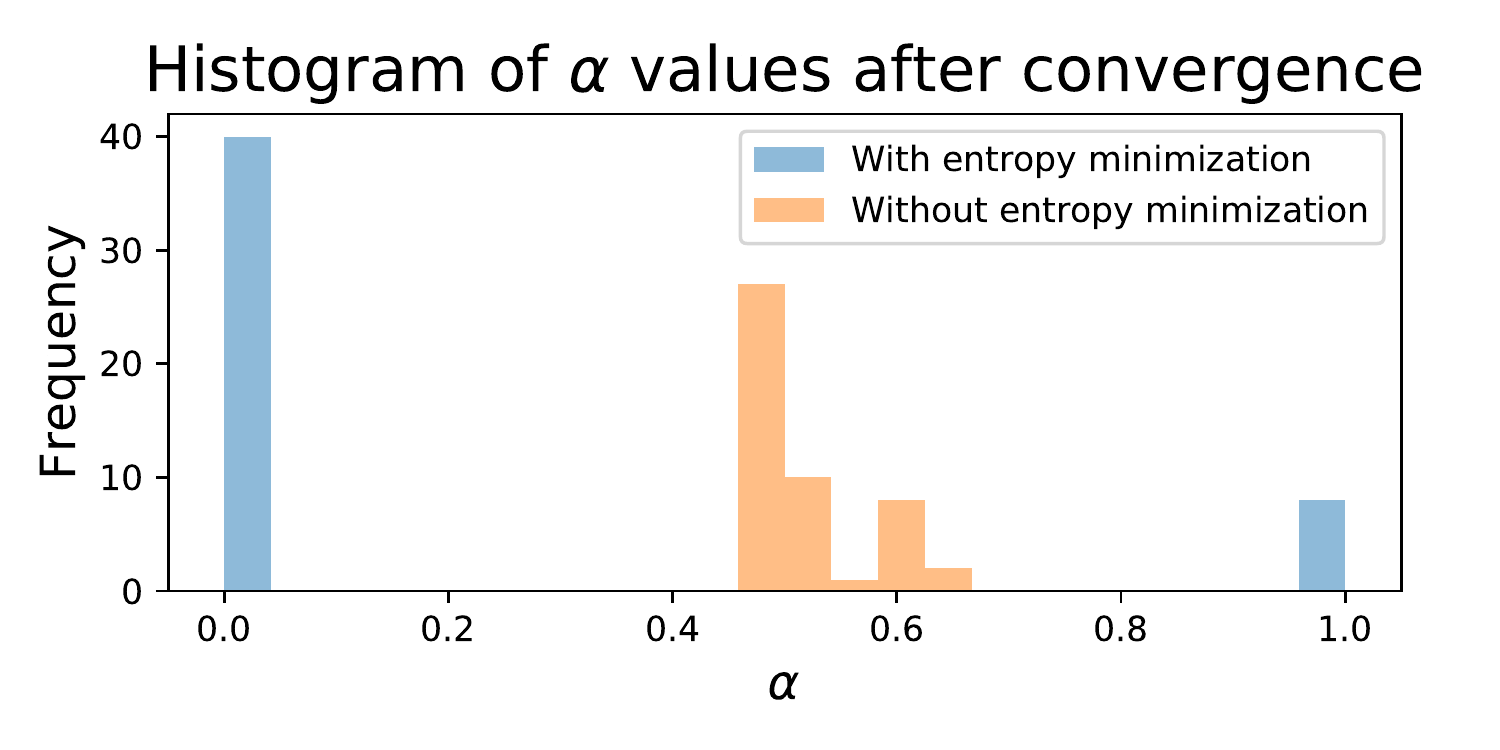}
\vspace{-5mm}
\caption{\label{fig:histo_alpha} The histogram of converged $\alpha$ from \emph{stochastic} continuous relaxation with and without \emph{minimum entropy regularization}. This shows that the minimum entropy regularization efficiently regularizes the distribution of $\alpha$'s, therefore leading to a smaller variance in sampling $\boldsymbol{X}$ by Eq. \eqref{eq:concrete}. We draw 25 bins uniformly from 0 to 1, where each bin represents an interval with 0.04 (best viewed in color).}
\vspace{-4mm}
\end{figure}

\begin{remark}
It is desirable to have explicit regularization on $\alpha_{ij}$ to avoid it from converging around $0.5$, so as to reduce the sampling variance. Interestingly, such a motivation also leads us to minimize the uncertainty of $\alpha_{ij}$. We implement this also by minimum entropy regularization on $\alpha_{ij}$.
\end{remark}

In summary, the proposed minimum entropy regularization alleviates both the objective bias of the deterministic methods and the sampling variance of the stochastic methods, which makes the search optimization converge to a single model that is ready for evaluation.

\subsection{Losses and Optimization} \label{sec:optimization}
After imposing the minimum entropy regularization, the full losses of our problem for both deterministic and stochastic versions are shown in Eqs. \eqref{eq:loss_deter} and \eqref{eq:loss_sto}:
\begin{align}
    \mathcal{L}_{\theta}(\boldsymbol{\alpha}) = \mathcal{L}_{\theta}^A(\boldsymbol{\alpha}) + \lambda \mathcal{L}_{\theta}^B(\boldsymbol{\alpha}) + \frac{\gamma}{n} \sum_{i,j} \mathcal{H}(\alpha_{ij}), \label{eq:loss_deter} \\
    \mathcal{L}_{\theta}(\boldsymbol{X}) = \mathcal{L}_{\theta}^A(\boldsymbol{X}) + \lambda \mathcal{L}_{\theta}^B(\boldsymbol{X}) + \frac{\gamma}{n} \sum_{i,j} \mathcal{H}(\alpha_{ij}), \label{eq:loss_sto}
\end{align}
where $\mathcal{H}(\alpha_{ij}) = - \alpha_{ij} \log \alpha_{ij} - (1 - \alpha_{ij}) \log (1 - \alpha_{ij})$ is the entropy of $\alpha_{ij}$, $\gamma$ is the regularization weight, and $\mathcal{L}_{\theta}^A$ and $\mathcal{L}_{\theta}^B$ are the losses for tasks $A$ and $B$. $\boldsymbol{X}$ for the stochastic method is sampled from $q_{\boldsymbol{\alpha}}(\boldsymbol{X})$ in Eq. \eqref{eq:concrete}.

Our optimization for $\boldsymbol{\alpha}$ and $\theta$ iterates the following steps:
\begin{enumerate}
    \item Sample two distinct batches of training data $\mathcal{X}_1=\{x_i\}_{i=1}^n$ with label $\mathcal{Y}_1=\{y_i\}_{i=1}^n$ and $\mathcal{X}_2=\{x_j\}_{j=1}^n$ with label $\mathcal{Y}_2=\{y_j\}_{j=1}^n$, where $\mathcal{X}_1\cap\mathcal{X}_2=\emptyset$.
    \item Compute the network output $\mathcal{O}_1$ of $\mathcal{X}_1$ with either deterministic Eq. \eqref{eq:darts} or stochastic Eq. \eqref{eq:reparameterized_computation} given current $\boldsymbol{\alpha}$ and $\theta$, and then update $\theta$ by $\nabla_{\theta} \mathcal{L}(\mathcal{O}_1, \mathcal{Y}_1)$.
    \item Compute the network output $\mathcal{O}_2$ of $\mathcal{X}_2$ with either deterministic Eq. \eqref{eq:darts} or stochastic Eq. \eqref{eq:reparameterized_computation} given current $\boldsymbol{\alpha}$ and $\theta$, and then update $\boldsymbol{\alpha}$ by $\nabla_{\boldsymbol{\alpha}} \mathcal{L}(\mathcal{O}_2, \mathcal{Y}_2)$.
\end{enumerate}

\begin{figure}[t]
\centering
\includegraphics[width=\linewidth]{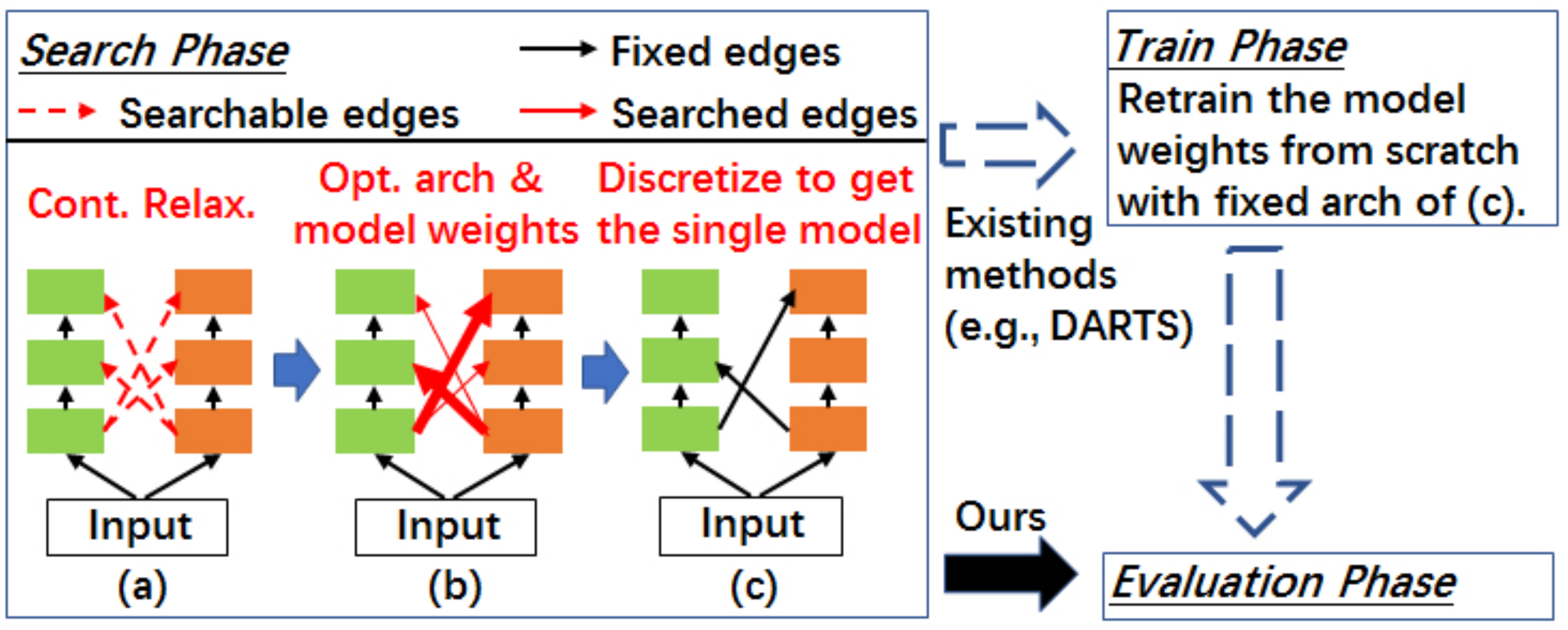}
\caption{\label{fig:search_algo} The overall procedure of the proposed search algorithm. The line width of the searched edges (solid red lines) indicate the converged architecture weights. \emph{Our method incorporates minimal entropy regularization on the architecture weights during the search phase, which makes the search optimization (b) converge closely to (c), therefore avoiding the train phase and achieving a better performance with less training time}. On the contrary, without architecture weights entropy minimization (\eg, DARTS), (b) and (c) can be very different, which requires retraining the model weights of (c) from scratch (best viewed in color).}
\end{figure}

\emph{Note that once the above iterations converge, the architecture, along with the model weights, can be used directly for evaluation, without the need of (re-)training the model weights from scratch}. This is because that the proposed minimal entropy regularization enables the search optimization to converge to a single network. The overall procedure of our method is shown in Fig. \ref{fig:search_algo}.

\noindent \textbf{Connections to DARTS and SNAS.} Our method alleviates the objective bias in DARTS and the sampling variance in SNAS. This is achieved by the unified, simple, and effective (verified in the ablation section) minimum entropy regularization on the architecture weight $\alpha_{ij}$. 

Our analysis also enables to re-incorporate the continuous relaxation methods and the discretization in a novel way, \eg, \emph{deterministic} continuous relaxation plus \emph{stochastic} discretization, or \emph{stochastic} continuous relaxation plus \emph{deterministic} discretization\footnote{Though we show in the Table \ref{Table:regu-A} that the different combinations make an insignificant performance difference, we hope that the analysis in this section sheds light on inventing novel NAS methods.}.

\section{Experiments}
In this section, we would like to investigate i) the \textbf{performance}, \ie, how the proposed MTL-NAS performs against the state-of-the-art \emph{general-purpose} multi-task networks, and ii) the \textbf{generalizability}, \ie, how the proposed MTL-NAS generalizes to various \emph{datasets}, \emph{network backbones}, and more importantly, various \emph{task sets}.

To validate the performance, we evaluate the proposed method against the state-of-the-art \textbf{NDDR-CNN} \cite{gao2019nddr} and \textbf{cross-stitch network} \cite{misra2016cross}. Additionally, we also provide various additional baselines for better evaluation: 

\noindent \textbf{Single-task Baseline}: it uses the single-task backbones.

\noindent \textbf{Multi-task Baselines}: it is the most intuitive multi-task network, which shares all the layers and splits at the last one. 

\noindent \textbf{MTL-NAS (SuperNet)}: it is the supernet of the proposed method before the NAS pruning, where all the intermediate layers from different tasks are connected, as long as they are connectable in the search space. This can also be regarded as a generalized version of the state-of-the-art NDDR-CNN. We leave its results in the Appendix (\ie, Table \ref{Table:VGG-16-A}).

\noindent \textbf{NDDR-CNN (SuperNet)}: this is a direct extension of NDDR-CNN analogous to MTL-NAS (SuperNet). The difference from MTL-NAS (SuperNet) is that we only allow interlinks between layers from the same CNN level. We leave its results in the Appendix (\ie, Table \ref{Table:VGG-16-A}). 

We also have extensive configurations to demonstrate the generalizability of our method:

\noindent \textbf{Datasets}: the NYU v2 \cite{silberman2012indoor} and the Taskonomy \cite{taskonomy2018}.

\noindent \textbf{Network Backbones}: VGG-16 \cite{simonyan2014very} and ResNet-50 \cite{he2016deep}.

\noindent \textbf{Task Sets}: \emph{pixel labeling tasks} including semantic segmentation and surface normal estimation, and \emph{image level tasks} including object classification and scene classification.

In the following, we first give the implementation details, and then present our results based on different task sets.

\subsection{Implementation Details}
For the VGG-16 backbone \cite{simonyan2014very}, we consider the features from all the convolutional layers. For ResNet-50 \cite{he2016deep}, we consider the features produced by each bottleneck block. We also constrain the search space due to the hardware limitation, \ie, we require the searchable \emph{task-opposite source} features (\ie, the source features from the opposite task from the target) to satisfy the following rules \emph{w.r.t} the fixed \emph{task-identical source} feature (\ie, the source features from the same task of the target): i) within the same stage, ii) from the same or earlier convolutional layers, and iii) no further by 3 layers apart. This produces search spaces with $2^{24}$ architectures for VGG-16, and $2^{37}$ for ResNet-50.

We perform 20000 training iterations for the VGG-16 backbone and 30000 iterations for the ResNet-50 backbone on the NYU v2 dataset \cite{silberman2012indoor}. On the Taxonomy dataset \cite{taskonomy2018} (3x larger than ImageNet), we train the network for $90000$ steps. We learn the model weights $\theta$ via an SGD optimizer with momentum $0.9$, weight decay $0.00025$, learning rate $0.001$ for VGG-16 backbone and $0.00025$ for ResNet-50 backbone. We use a poly learning rate decay with a power of $0.9$. We optimize architecture weights $\boldsymbol{\alpha}$ via an Adam optimizer \cite{kingma2014adam} with initial learning rate $0.003$ and weight decay $0.001$. We set $\gamma = 10$ for entropy minimization.

MTL-NAS is proxyless \emph{w.r.t.} both the \emph{datasets} and the \emph{architectures}, \ie, we directly search for the final architecture on the target (large) dataset. Our search time is only 12-13 and 27-28 GPU hours, respectively, for the NYUv2 experiments and the extremely large Taskonomy experiments, with the VGG-16 backbones on a single Nvidia Titan RTX GPU. The searched architectures and model weights can be used directly for evaluation without retraining from scratch.

\subsection{Semantic Segmentation and Surface Normal Estimation}
We use the NYU v2 dataset \cite{silberman2012indoor} for semantic segmentation and surface normal estimation. The groundtruth for 40 classes semantic labeling is from \cite{guptaCVPR13} and that for surface normal is precomputed from the depth map \cite{eigen2015predicting}. 

Per-pixel losses are used for both tasks. For semantic segmentation, we use the softmax cross-entropy loss, and calculate the pixel accuracy (PAcc), the mean intersection of union (mIoU) as the evaluation metrics. While for surface normal estimation, we train the network using the cosine loss (indicating the angle difference), and evaluate using Mean and Median angle distances of all the pixels, as well as the percentage of pixels that are within the angles of $11^\circ$, $22.5^\circ$, and $30^\circ$ to the ground-truth.

We perform this task on both VGG-16 and ResNet-50 network backbones, which are shown in Tables \ref{Table:VGG-16} and \ref{Table:Res-50}, respectively. The results show that our method outperforms state-of-the-art methods, demonstrating the effectiveness of the proposed method on semantic segmentation and surface normal estimation with different network backbones. 

\begin{table}[t]
\centering
\fontsize{8pt}{0.9\baselineskip}\selectfont
\begin{tabular}{@{\extracolsep{\fill}} l | c  c | c  c  c | c  c }
\hline
& \multicolumn{5}{c|}{\textbf{Surface Normal Prediction}} & \multicolumn{2}{c}{\textbf{Semantic Seg.}} \\
\hline
& \multicolumn{2}{c|}{\textbf{Err} \ ($\downarrow$)} & \multicolumn{3}{c|}{\textbf{Within $t^\circ$ (\%)} \ ($\uparrow$)} & \multicolumn{2}{c}{\textbf{(\%)} \ ($\uparrow$)} \\
\hline
& Mean & Med. & $11.25$ & $22.5$ & $30$ & mIoU & PAcc \\
\hline
Single  & 15.6 & 12.3 & 46.4 & 75.5 & 86.5 & 33.5 & 64.1 \\
Multiple   & 15.2 & 11.7 & 48.4 & 76.2 & 87.0 & 33.4 & 64.2 \\
C.-S.  & 15.2 & 11.7 & 48.6 & 76.0 & 86.5 & 34.8 & 65.0   \\
NDDR   & 13.9 & 10.2 & 53.5 & 79.5 & 88.8 & 36.2 & 66.4      \\
MTL-NAS   & \textbf{\underline{12.6}} & \textbf{\underline{8.9}}  & \textbf{\underline{59.1}} & \textbf{\underline{83.3}} & \textbf{\underline{91.2}} & \textbf{\underline{37.6}} & \textbf{\underline{67.9}} \\
\hline
\end{tabular}
\caption{Semantic segmentation and surface normal prediction on the NYU v2 dataset using the \textbf{VGG-16} network. C.-S. represents the the cross-stitch network. $\uparrow$/$\downarrow$ shows the higher/lower the better.}
\label{Table:VGG-16}
\vspace{-1.5mm}
\end{table}

\begin{table}[t]
\centering
\fontsize{8pt}{0.9\baselineskip}\selectfont
\begin{tabular}{@{\extracolsep{\fill}} l | c  c | c  c  c | c  c }
\hline
& \multicolumn{5}{c|}{\textbf{Surface Normal Prediction}} & \multicolumn{2}{c}{\textbf{Semantic Seg.}} \\
\hline
& \multicolumn{2}{c|}{\textbf{Err} \ ($\downarrow$)} & \multicolumn{3}{c|}{\textbf{Within $t^\circ$ (\%)} \ ($\uparrow$)} & \multicolumn{2}{c}{\textbf{(\%)} \ ($\uparrow$)} \\
\hline
& Mean & Med. & $11.25$ & $22.5$ & $30$ & mIoU & PAcc \\
\hline
Single  & 16.2 & 13.6 & 41.6 & 74.1 & 86.5 & 34.5 & 65.5 \\
Multiple   & 16.6 & 14.2 & 39.2 & 73.8 & 86.5 & 34.8 & 65.1 \\
C.-S.  & 16.6 & 14.3 & 39.1 & 73.7 & 86.5 & 34.8 & 65.7   \\
NDDR   & 16.4 & \textbf{\underline{12.8}} & 42.6 & 73.3 & \textbf{\underline{86.6}} & 36.7 & 66.7      \\
MTL-NAS   & \textbf{\underline{16.2}} & \textbf{\underline{12.8}} & \textbf{\underline{44.8}} & \textbf{\underline{73.9}} & 85.7 & \textbf{\underline{38.6}} & \textbf{\underline{68.6}} \\
\hline
\end{tabular}
\caption{Semantic segmentation and surface normal prediction on the NYU v2 dataset using the \textbf{ResNet-50} network. C.-S. represents the the cross-stitch network. $\uparrow$/$\downarrow$ shows the higher/lower the better.}
\label{Table:Res-50}
\vspace{-1.5mm}
\end{table}

\subsection{Object Classification and Scene Classification}
We evaluate object classification and scene classification tasks on the extremely large Taskonomy dataset \cite{taskonomy2018} (3x larger than ImageNet). 
We use the tiny split with data collected from 40 buildings. For both object and scene classifications, we use the  $\ell_2$ distance between model prediction and the soft class probabilities distilled from pretrained networks as the loss. We report the Top-1 and Top-5 recognition rates for both tasks.

The results are shown in Table \ref{Table:taskonomy}, which exhibit promising performance of our method on a different task set.

\begin{table}[t]
\centering
\scalebox{0.8}{
\begin{tabular}{@{\extracolsep{\fill}}  l | c  c | c  c }
\hline
& \multicolumn{2}{c|}{\textbf{Object}} & \multicolumn{2}{c}{\textbf{Scene}} \\
\hline
& \multicolumn{2}{c|}{\textbf{RecRate (\%)} \ ($\uparrow$)} & \multicolumn{2}{c}{\textbf{RecRate (\%)} \ ($\uparrow$)} \\
\hline
& Top 1 & Top 5  & Top 1  & Top 5  \\
\hline
Single & 33.8 & 63.0 & 37.8 & 70.5    \\
Multiple  & 34.1 & 66.1 & 37.8 & 71.2    \\
Cross-Stitch & 33.2 & 65.2 & 34.0 & 70.3    \\
NDDR  & 32.1 & 57.7 & 37.9 & 71.8    \\
MTL-NAS  & \textbf{\underline{34.8}} & \textbf{\underline{67.0}} & \textbf{\underline{38.2}} & \textbf{\underline{72.5}} \\
\hline
\end{tabular}}
\caption{Object classification and scene classification on the Taskonomy dataset using the VGG-16 network. $\uparrow$/$\downarrow$ shows the higher/lower the better.}
\label{Table:taskonomy}
\vspace{-2mm}
\end{table}

\section{Ablation Analysis}
In this section, we investigate the different choices of the building blocks of the proposed MTL-NAS by ablation analysis. Specifically, we are especially interested in the following questions: i) How does the proposed search algorithm perform \emph{w.r.t.} the baseline methods DARTS \cite{liu2018darts} and SNAS \cite{xie2018snas}? ii) How to initialize the novel inter-task layers into the fixed and well-trained backbones? iii) How to set the learning rate for the novel inter-task layers? 

We answer these questions in the following. We put the learning rate analysis in the Appendix (\ie, Table \ref{Table:lr}), which shows that the training of the novel architectures is not sensitive to learning rates. We also include the illustrations of the learned architectures into the Appendix (\ie, Fig \ref{fig:arch}), demonstrating that the learned architectures are \emph{heterogeneous and asymmetric}, which are arguably very difficult to be discovered by human experts. 

We perform all the ablation analyses using the VGG-16 network \cite{simonyan2014very} for semantic segmentation and surface normal estimation on the NYU v2 dataset \cite{silberman2012indoor}.

\subsection{Search Algorithm}

In this section, we take a closer investigation on the search algorithm, specifically, the \textbf{continuous relaxation}, the \textbf{discretization}, and most importantly, the \textbf{entropy minimization}. We validate the \emph{entropy minimization associating} with the \emph{deterministic} continuous relaxation and \emph{deterministic} discretization (analogous to DATRS w/o retraining), as well as the \emph{stochastic} continuous relaxation and \emph{stochastic} discretization (analogous to SNAS). We also provide the \emph{random search} baseline for comparison.

The experimental results are shown in Table \ref{Table:regu}, where both DARTS (w/o retraining, denoted as D in the table) and SNAS (denoted as S in the table) fail in our problem without minimal entropy regularization. We do not report the performance of SNAS due to the large sampling variance (see also Fig. \ref{fig:histo_alpha}). We also perform the stochastic method with minimal entropy regularization for 10 runs and witness negligible performance variance (see also Fig. \ref{fig:histo_alpha}). 

Moreover, it is interesting to witness that, after imposing the minimal entropy constraints, the deterministic and stochastic methods produces similar performances. We also perform different combinations of the continuous relaxation and discretization in the Appendix (\ie, Table \ref{Table:regu-A}), \ie, \emph{deterministic} continuous relaxation and \emph{stochastic} discretization, as well as the \emph{stochastic} continuous relaxation and \emph{deterministic} discretization. Those configurations also produce similar results (similar to DARTS plus MinEntropy and SNAS plus MinEntropy), which suggests the potential of our method to unify the popular DARTS and SNAS.

\begin{table}[t]
\centering
\fontsize{7pt}{0.9\baselineskip}\selectfont
\begin{tabular}{@{\extracolsep{\fill}} l l l | c  c | c  c  c | c  c }
\hline
& & & \multicolumn{5}{c|}{\textbf{Surface Normal Prediction}} & \multicolumn{2}{c}{\textbf{Semantic Seg.}} \\
\hline
& & & \multicolumn{2}{c|}{\textbf{Err} \ ($\downarrow$)} & \multicolumn{3}{c|}{\textbf{Within $t^\circ$ (\%)} \ ($\uparrow$)} & \multicolumn{2}{c}{\textbf{(\%)} \ ($\uparrow$)} \\
\hline
D & S & E & Mean & Med. & $11.25$ & $22.5$ & $30$ & mIoU & PAcc \\
\hline
\multicolumn{3}{@{} l|}{Random Search} & 14.1 & 10.1 & 53.9 & 79.2 & 88.2 & 35.3 & 66.1 \\
\hline
\checkmark             & & & 15.9 & 12.7 & 45.3 & 74.0 & 85.9 & 19.1 & 46.1 \\
 & \checkmark            & & - & - & - & - & - & - & - \\
\hline
\checkmark & & \checkmark  & 12.7 & \textbf{\underline{8.9}} & 58.9 & 83.1 & 90.9 & \textbf{\underline{37.7}} & \textbf{\underline{67.9}} \\
 & \checkmark & \checkmark & \textbf{\underline{12.6}} & \textbf{\underline{8.9}}  & \textbf{\underline{59.1}} & \textbf{\underline{83.3}} & \textbf{\underline{91.2}} & 37.6 & \textbf{\underline{67.9}} \\
 \hline
\end{tabular}
\caption{Effects of continuous relaxation, discretization, and entropy minimization. D denotes the \textbf{deterministic} method (\ie, DARTS without retraining), S means the \textbf{stochastic} method (\ie, SNAS), and E represents \textbf{minimum entropy regularization}. We do not report the results of the stochastic method as it produces too large sampling variances (see also the corresponding histogram of the converged architecture weights in Fig. \ref{fig:histo_alpha}). 
$\uparrow$/$\downarrow$ shows the higher/lower the better.}
\label{Table:regu}
\vspace{-1mm}
\end{table}



\subsection{Weight Initialization for the Novel Layers}
We are interested in investigating the initialized weights for the novel inter-task feature fusion layers. As we are inserting novel architectures into the fixed and well-trained single-task backbones, intuitively, we should make minimal changes of the original single-task output at each layer.

The ablation analysis is performed on different initializations of $w_t$, which is defined in Eq. \eqref{eq:fusion_op} and illustrated in the bottom right subfigure of Fig. \ref{fig:search_space}. The results shown in Table \ref{Table:init_weights} align our intuition, where initializing $w_t$ with a larger value, \eg, 0.9 or 1.0, produces the best performance.

\begin{table}[t]
\centering
\fontsize{8pt}{0.9\baselineskip}\selectfont
\begin{tabular}{@{\extracolsep{\fill}} l | c  c | c  c  c | c  c }
\hline
& \multicolumn{5}{c|}{\textbf{Surface Normal Prediction}} & \multicolumn{2}{c}{\textbf{Semantic Seg.}} \\
\hline
& \multicolumn{2}{c|}{\textbf{Err} \ ($\downarrow$)} & \multicolumn{3}{c|}{\textbf{Within $t^\circ$ (\%)} \ ($\uparrow$)} & \multicolumn{2}{c}{\textbf{(\%)} \ ($\uparrow$)} \\
\hline
Init $w_{\texttt{TI}}$ & Mean & Med. & $11.25$ & $22.5$ & $30$ & mIoU & PAcc \\
\hline
Random&16.7& 12.8 & 45.4 & 71.8 & 82.9 & 30.3 & 61.9 \\
0   & 16.9 & 12.9 & 45.1 & 71.3 & 82.4 & 30.1 & 61.8 \\
0.1 & 16.6 & 12.6 & 45.8 & 72.2 & 83.1 & 31.4 & 63.1 \\
0.2 & 14.4 & 10.6 & 52.2 & 78.2 & 87.8 & 33.8 & 65.2 \\
0.5 & 13.5 & 9.8  & 55.2 & 80.6 & 89.4 & 36.7 & 67.4 \\
0.8 & 12.9 & 9.2  & 57.6 & 82.6 & 90.7 & 37.0 & 67.6 \\
0.9 & \textbf{\underline{12.6}} & 9.0  & 58.8 & \textbf{\underline{83.3}} & \textbf{\underline{91.3}} & 37.2 & 67.4 \\
1.0 & \textbf{\underline{12.6}} & \textbf{\underline{8.9}}  & \textbf{\underline{59.1}} & \textbf{\underline{83.3}} & 91.2 & \textbf{\underline{37.6}} & \textbf{\underline{67.9}} \\
\hline
\end{tabular}
\caption{Effects of different initializations of the 1x1 convolution in the fusing operation. $w$ is defined in Eq. \eqref{eq:fusion_op}, also in the bottom right subfigure of Fig. \ref{fig:search_space}. $\uparrow$/$\downarrow$ shows the higher/lower the better.}
\label{Table:init_weights}
\vspace{-2.mm}
\end{table}

\section{Conclusion}
In this paper, we employed NAS for general-purpose multi-task learning (GP-MTL). We first disentangled GP-MTL into the task-specific backbone and inter-task feature fusion connections. We then focused on searching for a good inter-task feature fusion strategy within a task agnostic search space. We also proposed a novel search algorithm that is able to close the performance gap between search and evaluation. Our search algorithm also generalizes the popular single-shot gradient-based methods such as DARTS and SNAS. We conducted detailed ablation analysis to validate the effect of each proposed component. The extensive experiments demonstrate the promising performance and the desirable generalizability (to various datasets, task sets, and single-task backbones) of the proposed method.
~\\
\vspace{-1mm}
~\\
\noindent \textbf{Acknowledgments.} This work was partially supported by NSFC 61773295 and 61771201, NSF of Hubei Province 2019CFA037, and Guangdong R\&D key project of China 2019B010155001.

\renewcommand{\theequation}{A\arabic{equation}}
\renewcommand{\thefigure}{A\arabic{figure}}
\renewcommand{\thetable}{A\arabic{table}}
\renewcommand{\thesubsection}{A\arabic{subsection}}
\setcounter{equation}{0}
\setcounter{table}{0}
\setcounter{figure}{0}

\section*{Appendix}
This appendix complements our main text by giving more details on the following issues:
\begin{itemize}
    \item Sect. \ref{sect:baselines} shows the experiments with stronger baselines, \ie, NDDR-CNN (Supernet) and MTL-NAS (Supernet).
    \item Sect. \ref{sect:lr} shows the ablation analysis with different learning rates on the novel inter-task layers.
    \item Sect. \ref{sect:combinations} shows the ablation analysis with different combinations on continuous relaxation (\emph{stochastic} and \emph{deterministic}), discretization (\emph{stochastic} and \emph{deterministic}), and entropy minimization.
    \item Sect. \ref{sect:illu} gives illustrations of the learned architectures.
\end{itemize}

All the additional experiments and ablation analyses in this Appendix are performed on the NYU v2 \cite{silberman2012indoor} dataset for semantic segmentation and surface normal estimation.

\subsection{Experiments with Stronger Baselines \label{sect:baselines}}
We perform additional experiments with stronger baselines in this section. The results are shown in Table \ref{Table:VGG-16-A}, where NDDR-CNN (Supernet) adds inter-task edges between every pair of nodes \emph{at the same CNN level}, and MTL-NAS (Supernet) uses all the inter-task edges that are included in the search space. 

Table \ref{Table:VGG-16-A} shows that our method outperforms both the NDDR-CNN (Supernet), and more interestingly, the MTL-NAS (Supernet) which takes advantage of the whole search space without pruning. Those results illustrate the necessity of performing neural architecture search (NAS) on the supernet, therefore further demonstrating the promising performance of the proposed method.
\begin{table}[h]
\centering
\fontsize{7.5pt}{0.9\baselineskip}\selectfont
\begin{tabular}{@{\extracolsep{\fill}} l | c  c | c  c  c | c  c }
\hline
& \multicolumn{5}{c|}{\textbf{Surface Normal Prediction}} & \multicolumn{2}{c}{\textbf{Semantic Seg.}} \\
\hline
& \multicolumn{2}{c|}{\textbf{Err} \ ($\downarrow$)} & \multicolumn{3}{c|}{\textbf{Within $t^\circ (\%)$} \ ($\uparrow$)} & \multicolumn{2}{c}{\textbf{(\%)} \ ($\uparrow$)} \\
\hline
& Mean & Med. & $11.25$ & $22.5$ & $30$ & mIoU & PAcc \\
\hline
NDDR (S)    & 13.8 & 10.0 & 54.7 & 80.2 & 89.1 & 36.5 & 67.2 \\
MTL-NAS (S) & 13.3 & 9.5  & 56.4 & 81.4 & 89.9 & 36.1 & 66.9 \\
\hline
MTL-NAS   & \textbf{\underline{12.6}} & \textbf{\underline{8.9}}  & \textbf{\underline{59.1}} & \textbf{\underline{83.3}} & \textbf{\underline{91.2}} & \textbf{\underline{37.6}} & \textbf{\underline{67.9}} \\
\hline
\end{tabular}
\caption{Semantic segmentation and surface normal prediction on the NYU v2 dataset using the VGG-16 network. This table complements Table \ref{Table:VGG-16} by reporting more results using supernet. $\uparrow$/$\downarrow$ represents the higher/lower the better.}
\label{Table:VGG-16-A}
\end{table}

\subsection{Learning Rates for the Novel Inter-Task Architectures \label{sect:lr}}
We are interested in the learning rates for the novel architectures and weights because our method inserts novel interlinks into fixed and well-trained network backbone branches. 

In Table \ref{Table:lr}, we investigate the learning rates that are 1-1000 times of the base learning rate. The results show that the result differences among different learning rates are subtle, demonstrating that the proposed method is not sensitive to different learning rates to well train the novel inter-task architectures.

\begin{table}[h]
\centering
\fontsize{8pt}{0.9\baselineskip}\selectfont
\begin{tabular}{@{\extracolsep{\fill}} l | c  c | c  c  c | c  c }
\hline
& \multicolumn{5}{c|}{\textbf{Surface Normal Prediction}} & \multicolumn{2}{c}{\textbf{Semantic Seg.}} \\
\hline
& \multicolumn{2}{c|}{\textbf{Err} \ ($\downarrow$)} & \multicolumn{3}{c|}{\textbf{Within $t^\circ$ (\%)} \ ($\uparrow$)} & \multicolumn{2}{c}{\textbf{(\%)} \ ($\uparrow$)} \\
\hline
Scale & Mean & Med. & $11.25$ & $22.5$ & $30$ & mIoU & PAcc \\
\hline
1x    & 12.6 & 8.9 & 59.1 & 83.3 & 91.2 & \textbf{\underline{37.6}} & \textbf{\underline{67.9}} \\
10x   & \textbf{\underline{12.4}} & \textbf{\underline{8.8}} & \textbf{\underline{59.8}} & \textbf{\underline{84.0}} & \textbf{\underline{91.7}} & 37.5 & 67.8 \\
100x  & \textbf{\underline{12.6}} & \textbf{\underline{8.8}} & \textbf{\underline{59.8}} & 83.8 & 91.5 & 37.1 & 67.5 \\
1000x & 12.6 & 9.0 & 58.9 & 83.3 & 91.2 & 37.1 & 67.5 \\
\hline
\end{tabular}
\caption{Effects of different learning rates of the new searched layers. We investigate the learning rates that are 1, 10, 100, and 1000 times (\ie, scale) of the learning rate of the backbone networks, respectively.}
\label{Table:lr}
\vspace{-3mm}
\end{table}

\begin{figure*}[t]
\centering
\includegraphics[width=\linewidth]{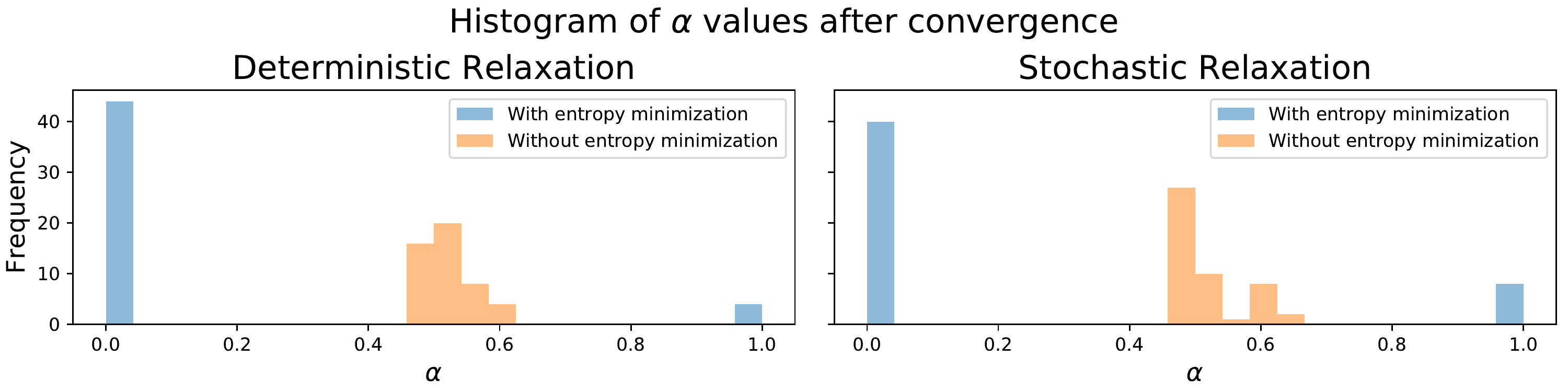}
\caption{\label{fig:histo_alpha} The histogram of converged $\alpha$ from \emph{deterministic} (Left) and \emph{stochastic} (Right, the same as Fig. 2 in the main paper) continuous relaxation with and without \emph{minimum entropy regularization}. This shows that the minimum entropy regularization efficiently regularizes the distribution of $\alpha$'s. We draw 25 bins uniformly from 0 to 1, where each bin represents an interval with 0.04.}
\end{figure*}

\begin{table*}[t]
\centering
\fontsize{9pt}{0.9\baselineskip}\selectfont
\begin{tabular}{@{\extracolsep{\fill}} c c c c c | c  c | c  c  c | c  c }
\hline
& & & & & \multicolumn{5}{c|}{\textbf{Surface Normal Prediction}} & \multicolumn{2}{c}{\textbf{Semantic Seg.}} \\
\hline
\multirow{2}{*}{\textbf{MinEntropy}} & \multicolumn{2}{c}{\textbf{Relaxation}} & \multicolumn{2}{c|}{\textbf{Discretization}}  & \multicolumn{2}{c|}{\textbf{Err} \ ($\downarrow$)} & \multicolumn{3}{c|}{\textbf{Within $t^\circ$ (\%)} \ ($\uparrow$)} & \multicolumn{2}{c}{\textbf{(\%)} \ ($\uparrow$)} \\
& D & S & D & S & Mean & Med. & $11.25$ & $22.5$ & $30$ & mIoU & PAcc \\
\hline
\multicolumn{5}{c|}{Random ~~~~~~~~~~Search} & 14.1 & 10.1 & 53.9 & 79.2 & 88.2 & 35.3 & 66.1 \\
\hline
& \checkmark & & \checkmark & & 15.9 & 12.7 & 45.3 & 74.0 & 85.9 & 19.1 & 46.1 \\
& \checkmark & & & \checkmark & - & - & - & - & - & - & - \\
& & \checkmark & \checkmark & & 14.7 & 11.1 & 50.4 & 77.3 & 87.7 & 29.4 & 60.5 \\
& & \checkmark & & \checkmark & - & - & - & - & - & - & - \\
\hline
\checkmark & \checkmark & & \checkmark &  & 12.7 & \textbf{\underline{8.9}} & 58.9 & 83.1 & 90.9 & 37.7 & \textbf{\underline{67.9}} \\
\checkmark & \checkmark & & & \checkmark & 12.7 & 9.0 & 58.7 & 82.9 & 90.9 & \textbf{\underline{37.9}} & \textbf{\underline{67.9}} \\
\checkmark & & \checkmark & \checkmark & & 13.1 & 9.3  & 57.0 & 81.7 & 90.1 & 37.3 & 67.7 \\
\checkmark & & \checkmark & & \checkmark & \textbf{\underline{12.6}} & \textbf{\underline{8.9}}  & \textbf{\underline{59.1}} & \textbf{\underline{83.3}} & \textbf{\underline{91.2}} & 37.6 & \textbf{\underline{67.9}} \\
\hline
\end{tabular}
\caption{Effects of different combinations of continuous relaxation (deterministic and stochastic), discretization (deterministic and stochastic), and entropy minimization. D denotes the \emph{deterministic}, S means the \emph{stochastic}, and MinEntropy represents minimum entropy regularization. Here, Relaxation (D) plus Discretization (D) without MinEntropy is equivalent to DARTS without re-training; Relaxation (S) plus Discretization (S) without MinEntropy is equivalent to SNAS. We did not report the results of methods relying on the Discretization (S) without MinEntropy as those methods produce too large sampling variances. 
$\uparrow$/$\downarrow$ represents the higher/lower the better.}
\label{Table:regu-A}
\end{table*}

\begin{figure*}[!htbp]
\centering
\includegraphics[width=0.8\linewidth]{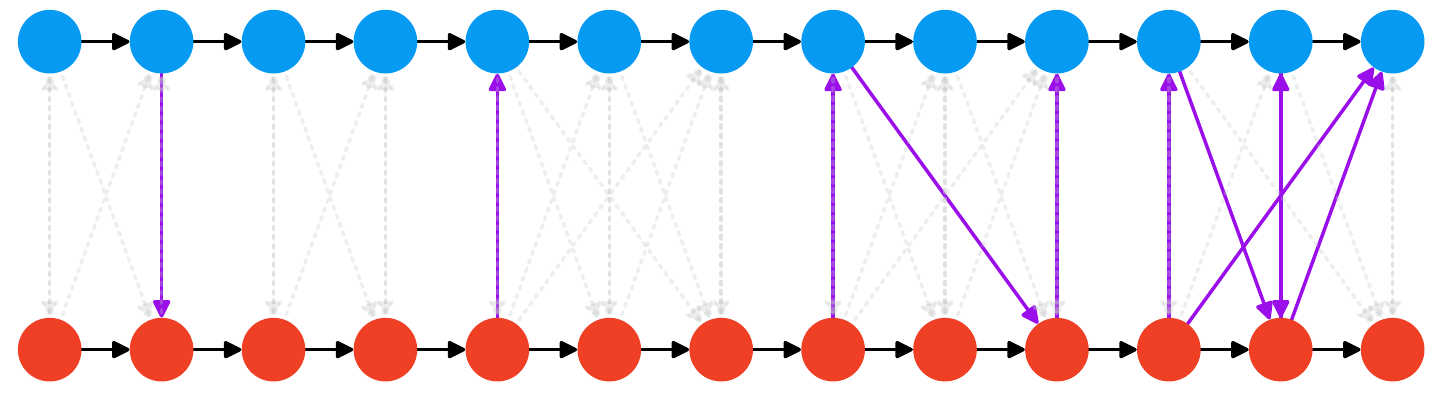}
\caption{\label{fig:arch} The learned architecture on VGG-16 backbones. Blue/Red nodes and the black solid arrows which link them represent the fixed VGG-16 single task backbones. The gray dash arrows are the candidate searchable edges for inter-task feature fusion (\ie, the search space defined by Sect. 5.1 in the main paper), and the purple solid arrows are the learned feature fusion edges.}
\end{figure*}

\subsection{Different Combinations of \emph{Deterministic} and \emph{Stochastic} Components in the Search Algorithm \label{sect:combinations}}
We complement our ablation analysis in Table 4 of the main text. Specifically, as shown in Sect. 4.3 (Connections to DARTS and SNAS) of the main text, by unifying the deterministic differentiable architecture search (DARTS) \cite{liu2018darts} and the  stochastic neural architecture search (SNAS) \cite{xie2018snas} into a more general single-shot gradient-based search algorithm framework, the proposed method is able to extend them by i) imposing entropy minimization to alleviate the objective bias in DARTS and the sampling variance in SNAS, and ii) enabling different combinations of the \emph{stochastic} and \emph{deterministic} components (\ie, the continuous relaxation and the discretization) as a result of i).

The results in Table \ref{Table:regu-A} show that i) imposing minimum entropy regularization significantly improves the performance for all combination cases of the continuous relaxation and the discretization; ii) \emph{under minimum entropy regularization}, different choices of discretization produce negligible performances when the same continuous relaxation method is used. This is because the uncertainty of $\alpha$ is significantly reduced with minimum entropy regularization, \ie, most $\alpha$'s converge to around 0's and 1's, making the sampling procedure of the stochastic discretization behave much more ``deterministic''. We show the histograms of the converged $\alpha$ under the deterministic or stochastic continuous relaxation, with or without minimum entropy regularization in Fig. \ref{fig:histo_alpha}, which demonstrate that most $\alpha$'s indeed converge to 0's and 1's with minimum entropy regularization\footnote{Note that the discretization procedure does not affect the convergence of $\alpha$'s, which performs a hard pruning (\ie, deterministic discretization) or a sampling (\ie, stochastic discretization) on the converged $\alpha$'s to deduce the final discrete architectures.}.

The phenomena discussed above further demonstrate that the proposed method successfully generalizes and improves the popular DARTS and SNAS methods, and the introduced unified framework is expected to shed light on designing novel single-shot gradient-based search algorithms.

\subsection{Illustrations of the Learned Architectures \label{sect:illu}}
We illustrate the learned architecture in Fig. \ref{fig:arch}, which demonstrates that the learned architecture is heterogeneous and asymmetric, therefore being arguably very difficult to be discovered by human experts.

\newpage
{\small
\bibliographystyle{ieee_fullname}
\bibliography{egbib}
}

\end{document}